\pdfoutput=1
\documentclass[11pt]{article}
\usepackage[final]{acl}

\usepackage{times}
\usepackage{latexsym}
\usepackage{algorithm}
\usepackage{algorithmic}
\usepackage{multirow}
\usepackage{enumitem}
\usepackage{xcolor}
\usepackage[T1]{fontenc}
\usepackage[utf8]{inputenc}
\usepackage{booktabs}

\usepackage{microtype}
\usepackage{inconsolata}
\usepackage{graphicx}

\usepackage{amssymb}
\usepackage{bm}
\usepackage{graphicx}
\usepackage{amsmath}
\usepackage{amsthm}
\usepackage{mathtools}
\usepackage{subcaption}
\usepackage{tcolorbox}
\usepackage{listings}
\usepackage{makecell}
\usepackage[normalem]{ulem}
\useunder{\uline}{\ul}{}
\usepackage{tabularx}

\title{FaithfulRAG: Fact-Level Conflict Modeling for Context-Faithful Retrieval-Augmented Generation}

\author{
Qinggang Zhang$^{1}$\thanks{Equal contribution.}, 
Zhishang Xiang$^{2}$\footnotemark[1], 
Yilin Xiao$^{3}$, 
Le Wang$^{4}$, 
Junhui Li$^{5}$,\\
{\bf Xinrun Wang}$^{6}$,
{\bf Jinsong Su}$^{1,7}$\thanks{Corresponding author.} \\
$^{1}$School of Informatics, Xiamen University \\
$^{2}$Institute of Artificial Intelligence, Xiamen University \\
$^{3}$The Hong Kong Polytechnic University \quad
$^{4}$Migu Meland Co., Ltd \quad
$^{5}$Soochow University \\
$^{6}$Singapore Management University \quad
$^{7}$Shanghai Artificial Intelligence Laboratory \\
\texttt{\{zqg.zhang, xzs.xiang\}@hotmail.com  jssu@xmu.edu.cn} 
}

\begin{document}
\maketitle
\begin{abstract}

Large language models (LLMs) augmented with retrieval systems have demonstrated significant potential in handling knowledge-intensive tasks. However, these models often struggle with unfaithfulness issues, generating outputs that either ignore the retrieved context or inconsistently blend it with the LLM's parametric knowledge. This issue is particularly severe in cases of knowledge conflict, where the retrieved context conflicts with the model’s parametric knowledge. While existing faithful RAG approaches enforce strict context adherence through well-designed prompts or modified decoding strategies, our analysis reveals a critical limitation: they achieve faithfulness by forcibly suppressing the model's parametric knowledge, which undermines the model’s internal knowledge structure and increases the risk of misinterpreting the context. To this end, this paper proposes FaithfulRAG, a novel framework that resolves knowledge conflicts by explicitly modeling discrepancies between the model’s parametric knowledge and retrieved context. Specifically, FaithfulRAG identifies conflicting knowledge at the fact level and designs a self-thinking process, allowing LLMs to reason about and integrate conflicting facts before generating responses. Extensive experiments demonstrate that our method outperforms state-of-the-art methods. The code is available at \textcolor{blue}{\url{https://github.com/DeepLearnXMU/Faithful-RAG}}.

\end{abstract}

\section{Introduction}
Large language models (LLMs), like GPT~\cite{openai2023gpt4}, Claude~\cite{anthropic2024claude} and DeepSeek series~\cite{liu2024deepseek}, have surprised the world with superior performance in many real-world applications~\cite{hong2024knowledge,yuan2025knapsack,zhou2024enhancing,zhou2025text}. Despite their effectiveness, LLMs are always criticized for their limited ability to handle knowledge-intensive tasks~\cite{huang2023survey,chen2024entity}, especially when faced with questions requiring professional or private knowledge not covered in their training corpus~\cite{zhang2024knowgpt}. Retrieval-augmented generation (RAG)~\cite{zhang2025survey,gao2023retrieval,cao2024retainingkeyinformationhigh,liu2024retrievalaugmentedmultimodalchainofthoughtsreasoning} offers a promising solution to customize LLMs for specific domains. Rather than retraining LLMs to incorporate new knowledge~\cite{zhang2023instruction} and updates~\cite{wang2024knowledge}, RAG enhances language models by leveraging external knowledge without modifying the model architecture or parameters. This approach enables LLMs to generate responses by leveraging not only their parametric knowledge (i.e., the knowledge embedded in its parameters from training) but also real-time retrieved domain-specific information, thereby providing more accurate and reliable answers. 



\begin{figure}
    \centering
    \includegraphics[width=1\linewidth]{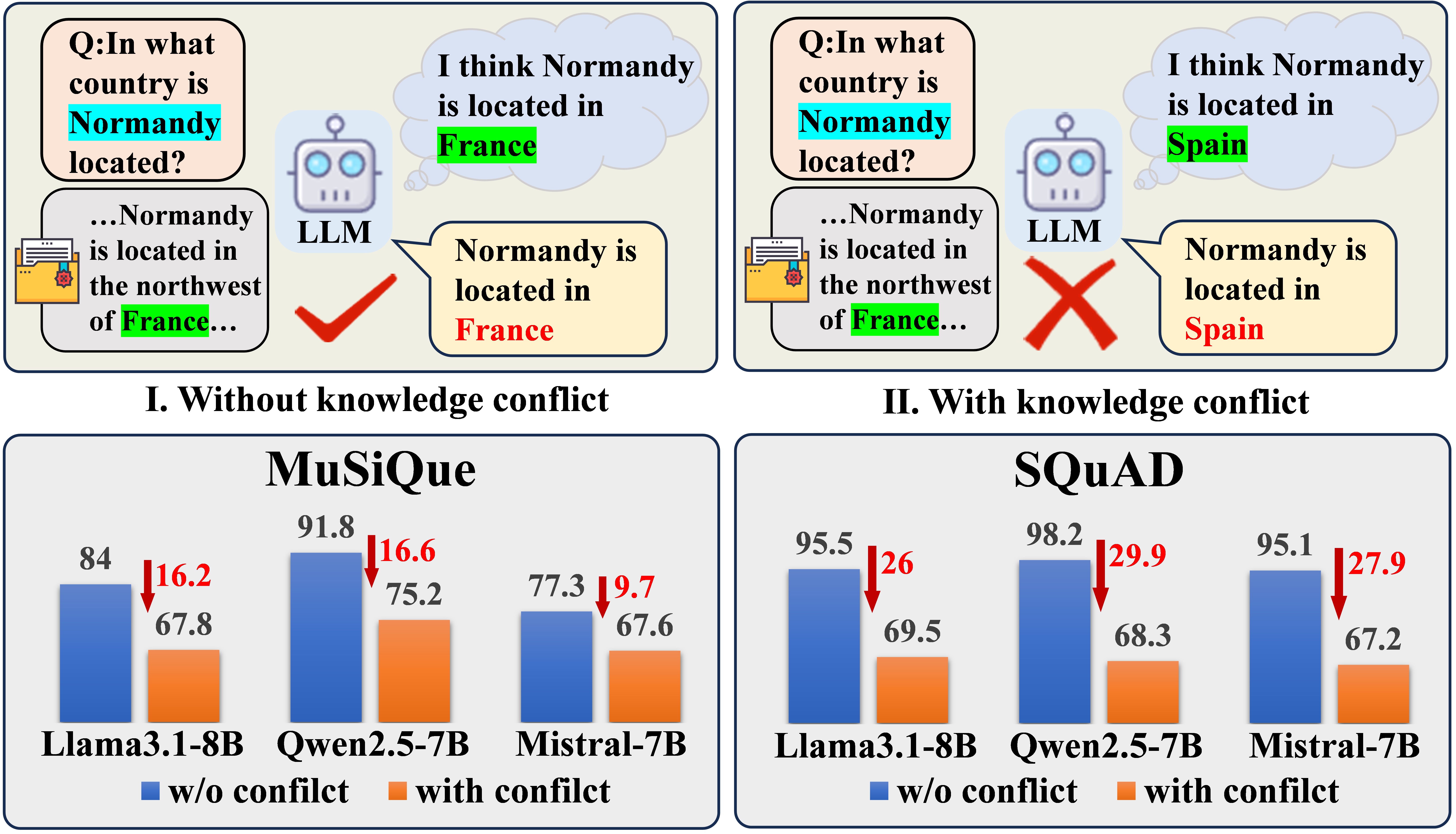}
    \vspace{-5mm}
    \caption{The running example of knowledge conflict and the performance comparison of different LLMs in scenarios with and without knowledge conflicts.
    }
    \label{fig:prior1}
    \vspace{-5mm}
\end{figure}


Despite recent advances, empirical studies~\cite{ming2025faitheval,coiced} have revealed that RAG systems struggle significantly in knowledge conflict scenarios~\cite{xu-etal-2024-knowledge-conflicts} where the retrieved context contradicts the parametric knowledge that the model has learned during pre-training. Such conflicts can lead to severe unfaithfulness issues~\cite{wu2024faithful},
where the model either generates a response that contradicts the context or fails to incorporate crucial details from the retrieved contexts. The results in Figure~\ref{fig:prior1} provide clear evidence of this issue that LLMs struggle to maintain faithfulness and correctness when faced with conflicting information.
Recently, a few studies~\cite{ying-etal-2024-intuitive,coiced} have been explored to improve the faithfulness of RAG systems. These works can be roughly categorized into two main directions: (i) Prompting-based methods that explore various prompting strategies to guide RAG systems to generate contextually faithful responses. By providing explicit instructions or few-shot examples, models can be directed to prioritize retrieved information over parametric knowledge~\cite{zhou-etal-2023-context,ying-etal-2024-intuitive}. 
(ii) Decoding-based models achieve faithfulness by forcing LLMs to align closely with the provided context during the decoding process by modifying the underlying generation mechanism through entropy-based constraints~\cite{coiced} or contrastive decoding~\cite{cad}.


 
\begin{figure}[t]
  \centering
  \includegraphics[width=0.5\textwidth]{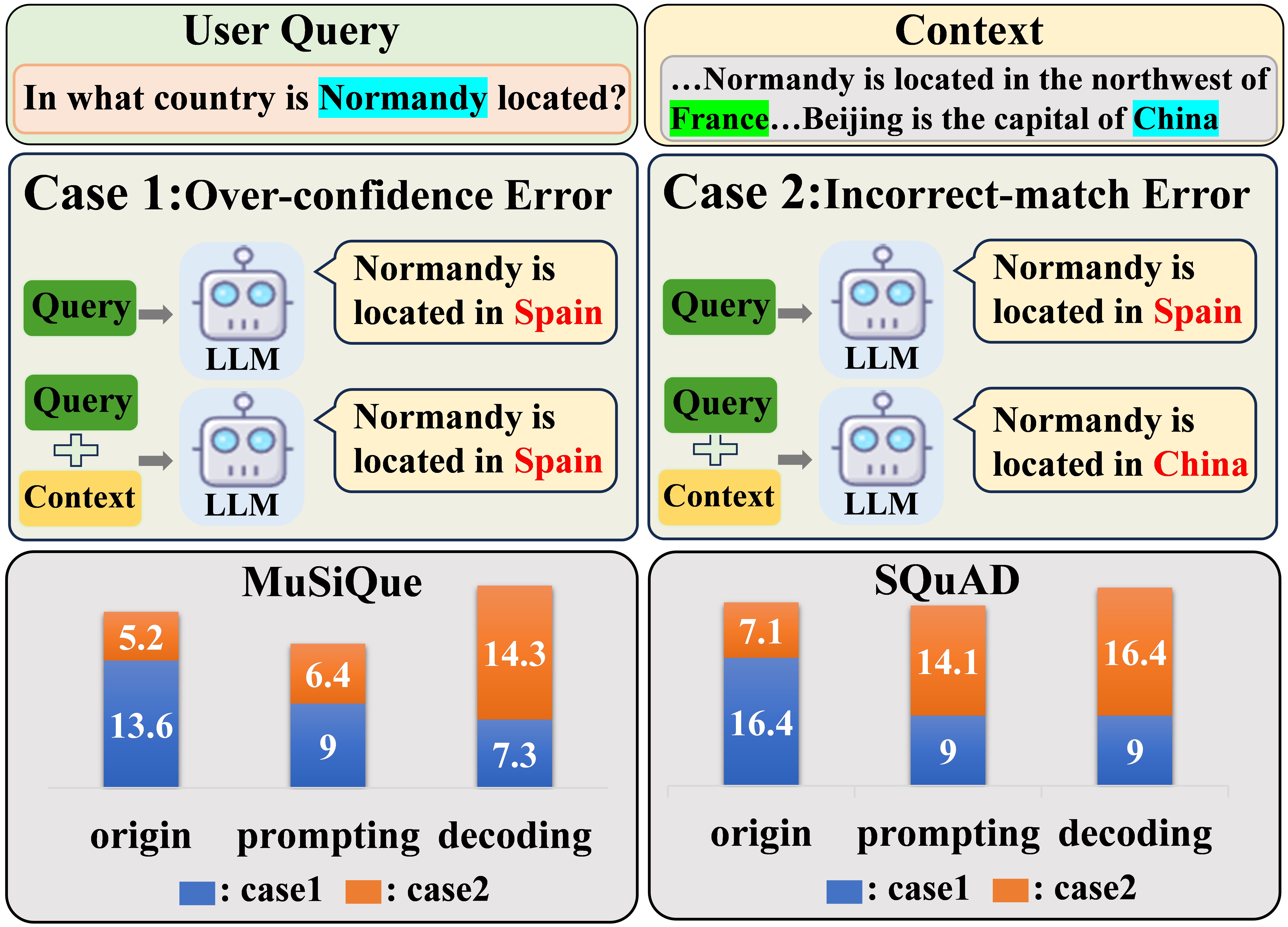} 
    \vspace{-7mm}
  \caption{The cases of errors and their distribution on MuSiQue and SQuAD datasets. The detailed experimental setup is introduced in Section~\ref{subsec:appendix-prior} of Appendix.}
  \label{fig:prior-data}
  \vspace{-6mm}
\end{figure}

However, we identified a critical limitation in existing faithful methods through a thorough analysis: while they can achieve context faithfulness, they often do so at the cost of increased risk of misinterpreting the context. As evidenced by our experimental results in Figure~\ref{fig:prior-data}, state-of-the-art context-faithful methods reduce unfaithful errors by 6.65\% but simultaneously cause a 6.42\% rise in incorrect match errors on average. This occurs because existing methods~\cite{zhou-etal-2023-context,ying-etal-2024-intuitive} attempt to achieve faithfulness by forcibly suppressing the model's confidence in its parametric knowledge without properly understanding and analyzing the differences between contextual information and the model's inherent knowledge. Such suppression compromises the model’s ability to critically evaluate and reconcile discrepancies, leading to degraded comprehension, logical incoherence, and a higher likelihood of aligning with incorrect contextual information.

In this paper, we propose a novel RAG model (FaithfulRAG) that achieves context faithfulness while maintaining accurate knowledge integration. Our research aims to address two fundamental research questions: (i) how to precisely locate conflicting knowledge between the model's parametric knowledge and the retrieved context, and (ii) how we could guide the language model to pay more attention to these critical segments during generation. 
Our major contribution is listed as follows:

\begin{itemize}
    \item We identify the key limitation of existing context-faithful methods and propose FaithfulRAG to improve the faithfulness of RAG systems without sacrificing accuracy.
    \item FaithfulRAG adopts a novel self-fact mining module to externalize the LLM's understanding of the question, obtaining fine-grained knowledge at the fact level.
    \item 
    FaithfulRAG identifies conflicting knowledge by aligning self-fact with contexts and designs a self-thinking module, allowing LLMs to reason about and integrate conflicting facts before generating responses. 
    \item Experiments show that FaithfulRAG outperforms state-of-the-art models, achieving more accurate and faithful generation. 
\end{itemize}

\section{Problem Statement}
Retrieval-augmented generation (RAG) systems consist of two key stages~\cite{RAG}: (i) \textbf{Knowldge Retrieval}, in which the model identifies and extracts semantically relevant documents from external knowledge sources based on the user query, which serves as the contextual foundation for subsequent processing.
(ii) \textbf{Generation Stage}: The retrieved context is synthesized with the user query to construct an augmented prompt, which LLMs process to generate a contextually grounded and factually consistent output.

In our work, we focus on the generation stage and guide LLM to generate more faithful responses by accurately interpreting the context with the model’s parametric knowledge.
\begin{equation}
P(a \mid Q, \mathcal{C}) = \prod_{t=1}^{T} P(a_t \mid a_{<t}, Q, \mathcal{C}; \theta)
\end{equation}
While $Q$ represents the user query, $\mathcal{C}$ represents the context, and $a$ is the target answer. $t$ denotes the position of the currently generated token, $T$ refers to the total number of tokens in the $a$, $\theta$ represents the parameters of the generator model.

\section{Preliminary Study}
Before going into the technique details of FaithfulRAG, we first conduct a preliminary study to identify the primary limitation of existing methods.

\begin{figure*}[t]
  \centering
  \includegraphics[width=1\textwidth]{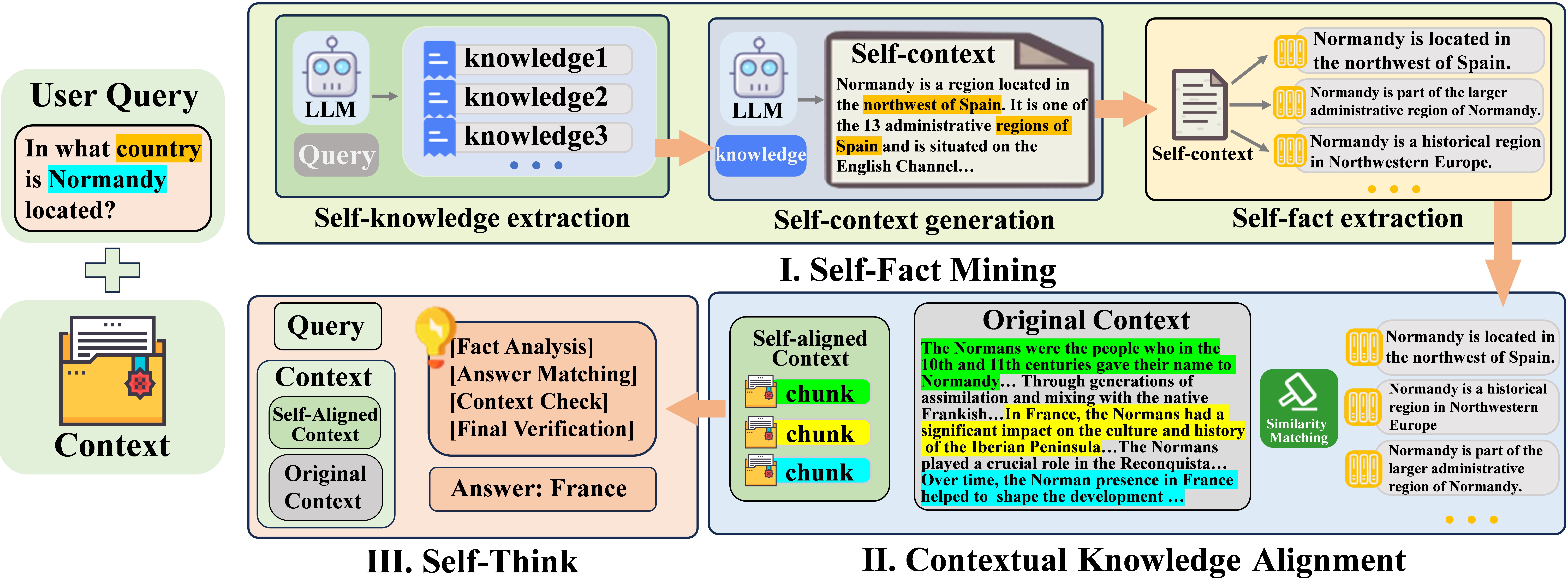} 
  \vspace{-5mm}
  \caption{The overall pipeline of our FaithfulRAG framework. FaithfulRAG first designs a self-fact mining module to externalize the LLM's understanding of the question, obtaining fine-grained knowledge at the fact level. Then it identifies conflicting knowledge by aligning self-fact with the retrieved contexts and adopts a self-thinking module, allowing LLMs to reason about and integrate conflicting facts before generating responses. }
  \vspace{-5mm}
  \label{fig:pipeline}
\end{figure*}

\paragraph{Performance Degradation in Knowledge Conflict Scenarios.} 
RAG systems often struggle with severe unfaithfulness issues, especially in knowledge conflict scenarios where retrieved contexts contradict the model's parametric knowledge~\cite{longpre-etal-2021-entity,zhou-etal-2023-context,xu-etal-2024-knowledge-conflicts}. As shown in Figure~\ref{fig:prior1}, our experiments on two real-world datasets reveal that LLMs always illustrate a significant performance gap between scenarios with and without knowledge conflicts. Specifically, we evaluate the performance of Llama3.1-8b-instruct, Qwen2.5-7b-instruct, and Mistral-7b-instruct on the MuSiQue and SQuAD datasets~\cite{ying-etal-2024-intuitive}. Without knowledge conflicts, Qwen2.5-7b-instruct achieves the highest factual accuracy, followed by Llama3-8b-instruct, while Mistral-7b-instruct has the lowest scores. However, when knowledge conflict is introduced, all models experience a significant drop in performance, with drops ranging from 9.7\% to 29.9\%.


\paragraph{Error Analysis.}\label{content:prior}

Through an in-depth analysis of the erroneous outputs in knowledge conflict scenarios, we found that the performance degradation is attributed to two dominant failure modes:

\noindent\textbf{Case 1: Over-confidence Error.} The LLM insists on its inherent parametric knowledge while ignoring the facts in the context, leading to unfaithful responses~\cite{ying-etal-2024-intuitive,xie2024adaptive}.

\noindent\textbf{Case 2: Incorrect-match Error.} The LLM updates its parametric knowledge but incorrectly learns from the misleading context.

\noindent To explore the underlying reasons, we evaluate state-of-the-art faithful RAG methods on the MuSiQue and SQuAD benchmarks~\cite{ying-etal-2024-intuitive} and present the error distribution in Figure~\ref{fig:prior-data}. The results reveal a critical limitation of existing models that they improve faithfulness at the cost of increasing the risk of misinterpreting the context. Specifically, we have the following observations:

\noindent\textbf{Obs. 1.} The vanilla RAG model (origin) exhibits a high proportion of Case 1 errors, with 13.6\% on MuSiQue and 16.4\% on SQuAD, while Case 2 errors remain relatively low at 5.2\% and 7.1\%, respectively. This suggests that the model primarily suffers from LLMs’ inherent bias toward their parametric knowledge. Such a tendency aligns with prior findings~\cite{xie2024adaptive} that LLMs tend to prioritize pre-trained knowledge over contextual evidence when faced with conflicting information.  


\noindent\textbf{Obs. 2.} The prompting-based faithful RAG model (prompting) reduces Case 1 errors significantly, from 13.6\% to 9\% in MuSiQue and from 16.4\% to 9\% in SQuAD. However, this comes at the cost of a substantial increase in Case 2 errors, which rise from 5.2\% to 6.4\% in MuSiQue and from 7.1\% to 14.1\% in SQuAD, reflecting an overcorrection where the model prioritizes context faithfulness without adequately validating the relevance or accuracy of the retrieved information.

\noindent\textbf{Obs. 3.} The decoding-based faithful RAG model (decoding) further decreases Case 1 errors, reducing them to 7.3\% in MuSiQue and 9\% in SQuAD. Compared to prompting-based approaches, it is more effective at mitigating over-confidence issues. However, this improvement comes with an even greater rise in Case 2 errors, which jump to 14.3\% in MuSiQue and 16.4\% in SQuAD. This suggests that the decoding-based model imposes stronger constraints on faithfulness but at the expense of a heightened risk of misinterpreting context.

\paragraph{Discussion.}
Existing faithful RAG models~\cite{zhou-etal-2023-context,coiced} reduce over-confidence errors but simultaneously cause a significant rise in incorrect match errors. This occurs because existing methods attempt to achieve faithfulness by forcibly suppressing the model's confidence in its parametric knowledge without properly understanding and analyzing the discrepancies between context and the model's inherent knowledge. 
Such suppression weakens the model’s ability to critically evaluate interactions between parametric knowledge and contextual evidence, leading to either rigid adherence to outdated parametric knowledge or 
severe contextual overfitting, where the model passively adopts retrieved claims without critically evaluating their correctness. 

Achieving true faithfulness requires identifying specific conflicting facts rather than broadly suppressing knowledge. By prioritizing why conflicts occur and how to resolve them at the fine-grained fact level, FaithfulRAG advances beyond brute-force suppression, enabling LLMs to act as critical, context-aware reasoners rather than passive knowledge retrievers, ensuring outputs are both contextually aligned and logically consistent.
\section{Methodology}
In this section, we introduce our FaithfulRAG in detail. As shown in Figure~\ref{fig:pipeline}, FaithfulRAG consists of three main components: (i) \textbf{Self-Fact Mining}, which is used to extract self-facts (query-related knowledge) by externalizing the LLM's parametric knowledge. (ii) \textbf{Contextual Knowledge Alignment}, that leverages self-facts to identify the most relevant information from the provided context. and (iii) \textbf{Self-Think}, enables the model to handle discrepancies between the context and its parametric knowledge at the fact level.

\subsection{Self-Fact Mining}\label{content:knowledge-externalization}

To identify conflicts between a model's stored knowledge and the contextual information it receives, we need a clear representation of the LLM's internal understanding of the problem at hand. This involves not only capturing the factual content the model possesses but also revealing the logical structure it uses to organize these facts. Our approach addresses both aspects by employing a hierarchical, three-stage process that externalizes the model's internal knowledge into distinct logical and fine-grained factual representations.

Specifically, for a given question $Q$, our objective is to extract the essential factual and conceptual information required to answer $Q$ while filtering out unnecessary details. We achieve this through the following three sequential stages: Self-Knowledge Extraction, Self-Context Generation and Self-Fact Extraction.




\paragraph{Self-Knowledge Extraction.} It surfaces the LLM’s implicit logical structure by identifying required knowledge domains and their interdependencies. Specifically, we prompt the LLM to identify the conceptual and factual prerequisites for answering $Q$, yielding a set of abstract, high-level knowledge:
\begin{equation}
    \mathcal{K}_{\text{self}}(Q) = \{k_1, k_2, \dots, k_n\},
\end{equation}
where $\mathcal{K}_{\text{self}}(Q)$ represents the extracted self-knowledge and where $k_i$ represents abstract claims.

\paragraph{Self-Context Generation.} Then, we translate abstract conceptual mappings into concrete narratives, ensuring alignment between high-level reasoning and specific factual claims. Specifically, the model synthesizes the abstract claims into a coherent narrative that contextualizes the target question. In contrast to previous works~\cite{tan2024blinded,yu2023generate}, which often generate context without any grounding, we explicitly condition the context generation on $\{k_1, k_2, \dots, k_n\}$. This allows for more coherent and logically consistent context, while ensuring relevance. Formally, the self-context is generated via a generator $\mathcal{G}_{1}$ as follows:
\begin{equation}
    \mathcal{C}_{\text{self}}(Q)=\mathcal{G}_{1}(Q,\mathcal{K}_{\text{self}}(Q)).
\end{equation}

\paragraph{Self-Fact Extraction.} 
The self-context is distilled into concrete factual assertions using the LLM as a fact extractor:
\begin{equation}
    \mathcal{F}_{\text{self}}(\mathcal{C}_{\text{self}}) = \{f_1, f_2, \dots, f_m\}.
\end{equation}
These self-facts serve as anchors for aligning with context while preserving logical constraints. By decoupling what the model knows from how it organizes knowledge, the framework supports differentiated error diagnosis and targeted corrections.

\subsection{Contextual Knowledge Alignment}

To resolve knowledge conflicts while preserving logical coherence, in this section, FaithfulRAG aligns the model’s externalized self-facts with the retrieved context through a structured, interpretable process. 
The alignment is structured as follows:

\noindent\textbf{Context Chunking}: The original context $\mathcal{C}_{\text{orig}}$ is divided into a set of chunks $\mathcal{C}_{\text{orig}}^{i}$, where $i=1, 2, \dots, m$. Smaller chunks allow granular comparison with self-facts, reducing noise from irrelevant text spans. Formally:
\begin{equation}
    \mathcal{C}_{\mathrm{orig}} = \bigcup_{i=1}^{m} \mathcal{C}_{\text{orig}}^{i}.
\end{equation}

\noindent\textbf{Similarity Matching}: We first embed self-facts $f_i$ extracted from the LLM's parametric knowledge and the chunks $\mathcal{C}_j$ divided from the context into a shared semantic space, and then measure their semantic distance by using cosine similarity:
\begin{equation}
    Sim(\textbf{f}_{i}, \textbf{c}_{j}) = \text{cos}(\textbf{f}_{i}, \textbf{c}_{j}).
\end{equation}

\noindent Then, we select the chunks $\mathcal{C}_{\text{aligned}}$ based on similarity scores, where $\mathcal{C}_{\text{aligned}}$ represents the chunks that are highly semantically aligned with self-facts that extracted from LLM's parametric knowledge.

    



 \subsection{Self-Think}
To resolve knowledge conflicts while ensuring contextual faithfulness, FaithfulRAG employs a Self-Think module  that dynamically synthesizes insights from two sources: 
(i) the self-aligned context \(\mathcal{C}_{\text{aligned}}\) (context segments conflicting or aligning with parametric knowledge) and 
(ii) the original context \(\mathcal{C}_{\text{orig}}\). 
This iterative workflow ensures the model critically evaluates discrepancies, mitigates overconfidence, and integrates evidence transparently.
We formalize this procedure as a cognitive function \(\mathcal{R}_{\text{STR}}\), which synthesizes key insights by comparing and merging relevant information from both contexts. Let the answer be generated as:
\begin{equation}
    \text{Answer} \;=\; \mathcal{R}_{\text{STR}}\bigl(\mathcal{C}_{\text{aligned}}, \mathcal{C}_{\text{orig}}\bigr).
\end{equation}

Specifically, this process can be divided into two parts: thinking and reasoning.  In the thinking stage, LLM first produces an initial answer from $\mathcal{C}_{\text{aligned}}$. It subsequently evaluates the reliability of this answer and ascertains whether $\mathcal{C}_{\text{aligned}}$ provides sufficient information. If the answer is found to be unreliable or incomplete, the model selectively incorporates relevant elements from $\mathcal{C}_{\text{orig}}$ to enrich the aligned context, thereby creating a fused context, $\mathcal{C}_{\text{fused}}$. This operation is defined as:

\begin{equation}
   \mathcal{C}_{\text{fused}}  = \mathcal{G}_{2}(\mathcal{C}_{\text{aligned}}, \mathcal{C}_{\text{orig}}),
\end{equation}
where $\mathcal{G}_{2}$ represents the context fusion function. 
In the reasoning module, the final answer is regenerated from $\mathcal{C}_{\text{fused}}$ through a step-by-step reasoning procedure, ensuring that the answer is both coherent and adequately supported by the evidence.

To make it more clear, the prompt templates applied in FaithfulRAG for self-knowledge extraction, self-context generation, self-fact extraction, and self-think are shown in Figure \ref{fig:prompts} of Appendix.

\label{content:Pre-Generation Perception}



\begin{table*}[t]
\centering
\caption{The comparison of performance between our model and SOTA baselines on four datasets. The best result for each dataset is highlighted in \textbf{bold}, while the best result for each backbone model is indicated with an \underline{underline}.}
\resizebox{.87\linewidth}{!}{
\begin{tabular}{cc|cccc}
\midrule
 \multirow{2}{*}{Model} & \multirow{2}{*}{Backbone LLM} & \multicolumn{4}{c}{Dataset} \\ 
 \cmidrule(lr){3-6}
 & &FaithEval & RealtimeQA &MuSiQue & SQuAD \\ 
\midrule
\multicolumn{6}{c}{\textbf{Group 1: Default Methods}} \\ \cmidrule(lr){1-6}
\multirow{2}[6]{*}{Origin model without context} & llama3.1-8b-instruct & 7.6 & 28.3 & 11.2 & 11.2\\
& qwen2.5-7b-instruct  &4.2 &40.7 &19.6 &11.1 \\ 
 & mistral-7b-instruct  &6.3 &29.2 &13.8 &11.5 \\ 
 \cmidrule(lr){2-6}
\multirow{2}[6]{*}{Origin model with full context} &  llama3.1-8b-instruct &63.3 &67.3 &67.8 & 69.5\\
 & qwen2.5-7b-instruct  &53.1 &78.7 &75.2 & 68.3 \\ 
 & mistral-7b-instruct  &61.9 &52.2 &67.6& 67.2 \\
\cmidrule(lr){1-6}

\multicolumn{6}{c}{\textbf{Group 2: Specific RAG Models}} \\ \cmidrule(lr){1-6}
Self-RAG~\cite{asai2023selfraglearningretrievegenerate} & Llama2-7B & 37.4 &55.8& 54.1 &62.0\\
ChatQA-1.5~\cite{liu2024chatqa} & Llama3.1-8B &56.2& 56.7& 75.0&  77.0 \\
ChatQA-2.0~\cite{xu2024chatqa} & Llama3.1-8B & 65.2& 57.5 &77.2 &75.4 \\
\cmidrule(lr){1-6}

\multicolumn{6}{c}{\textbf{Group 3: Context-faithful Prompting}} \\ \cmidrule(lr){1-6}
\multirow{2}[6]{*}{Opin(Instr)~\cite{zhou-etal-2023-context}} &llama3.1-8b-instruct & 68.1 &75.2& 70.3 &73.4 \\
&qwen2.5-7b-instruct & 56.7 &81.4& 76.9 & 70.5 \\
&mistral-7b-instruct & 62.5 & 51.3& 68.1 & 69.3 \\
\cmidrule(lr){2-6}
\multirow{2}[6]{*}{ATTR~\cite{zhou-etal-2023-context}} &llama3.1-8b-instruct & 63.8  &76.9 &62.8 & 69.5\\
&qwen2.5-7b-instruct & 58.1 &83.0 &\underline{78.7}&72.9  \\
&mistral-7b-instruct & 63.6 &52.2 &66.1 &70.2 \\
\cmidrule(lr){2-6}
\multirow{2}[6]{*}{KRE~\cite{ying-etal-2024-intuitive}} &llama3.1-8b-instruct & 51.6 &48.6$^*$& 35.9$^*$ &66.1 \\
&qwen2.5-7b-instruct & 59.6 &\textbf{\underline{86.7}}& 70.7 & 73.7 \\
&mistral-7b-instruct & 73.2 & 76.9& 50.6$^*$ & 74.6 \\

\cmidrule(lr){1-6}

\multicolumn{6}{c}{\textbf{Group 4: Context-faithful Decoding}} \\ \cmidrule(lr){1-6}
\multirow{2}[6]{*}{CAD~\cite{cad}} &llama3.1-8b-instruct & 66.2 &61.9 &72.6& 71.2 \\
&qwen2.5-7b-instruct & 60.5 &77.0 &78.6 & 73.4\\
&mistral-7b-instruct & 60.2 &55.8 &63.6 & 66.9  \\
\cmidrule(lr){2-6}
\multirow{2}[6]{*}{COIECD~\cite{coiced}} &llama3.1-8b-instruct & 67.7 &62.8 &70.5 & 71.8\\
&qwen2.5-7b-instruct & 62.3 &78.8 &69.7 & 70.8 \\
&mistral-7b-instruct & 62.8 &58.4 &66.8 & 65.4  \\

\cmidrule(lr){1-6}
\multirow{2}[6]{*}{FaithfulRAG (Ours)} &llama3.1-8b-instruct & \underline{79.8} &\underline{81.4} & \textbf{\underline{79.9}} & \textbf{\underline{86.3}}\\
&qwen2.5-7b-instruct & \underline{71.8} &84.1 &78.0 &\underline{78.3} \\
&mistral-7b-instruct & \textbf{\underline{81.7}} &\underline{77.0} &\underline{78.5}&\underline{85.7} \\
\bottomrule
\end{tabular}
}\\
\footnotesize{$^*$ There is a sharp decline in performance as the model refuses to generate responses. }
\label{tab:main-experient}
\end{table*}

\section{Experiment}
In this section, we conduct comprehensive experiments to verify the effectiveness of FaithfulRAG. Specifically, we aim to answer the following questions.  \textbf{Q1 (Effectiveness):} How does FaithfulRAG perform compared with SOTA competitors? \textbf{Q2 (Error analysis):} How effective is FaithfulRAG in alleviating different types of errors? \textbf{Q3 (Ablation study:)} How does each component of FaithfulRAG contribute to the performance? \textbf{Q4 (Case study):} How does it works in real-world scenarios? (Note that \textbf{Q4} is studied in Appendix~\ref{appendix:case}, while the first three questions are explored in the main content.)
\subsection{Experiment Settings}
\textbf{Datasets:} We evaluate FaithfulRAG on four benchmark datasets. \textbf{MuSiQue}~\cite{trivedi2022MuSiQuemultihopquestionssinglehop} and \textbf{SQuAD}~\cite{rajpurkar2016SQuAD} are from KRE~\cite{ying-etal-2024-intuitive} which introduce fact-level knowledge conflicts, where only contradictory factual statements appear in the context. In contrast, \textbf{FaithEval}~\cite{ming2025faitheval} introduces logical-level conflicts, where inconsistencies arise not from direct factual contradictions but from reasoning chains that lead to conflicting conclusions. We also include \textbf{RealtimeQA}~\cite{kasai2024realtime} dataset to test the model performance in extreme cases where some contexts are irrelevant to the question. More details of datasets can be found in Appendix~\ref{subsec:appendix-dataset}.

\noindent \textbf{Baselines:} We carefully select baselines from four categories for a comprehensive evaluation. \textbf{Default Methods}:  origin model without context, origin model with full context; \textbf{RAG models}: Self-RAG~\cite{asai2023selfraglearningretrievegenerate}, ChatQA-1.5~\cite{liu2024chatqa}, ChatQA-2.0~\cite{xu2024chatqa}; \textbf{Context-faithful Prompting}: Opin~\cite{zhou-etal-2023-context}, KRE~\cite{ying-etal-2024-intuitive}; and \textbf{Context-faithful Decoding}: CAD~\cite{cad}, COIECD~\cite{coiced}. More details are described in Section~\ref{subsec:appendix-baselines} of the Appendix. 
Additionally, we did not include SFR-RAG~\cite{nguyen2024sfr} as a baseline since its parameters have not been released.

\noindent \textbf{Evaluation Metrics and Implementation Details:} 
Following previous studies, we evaluate all models using accuracy (ACC), where a model’s response is considered correct only if it contains the ground truth answer. For the MuSiQue and SQuAD, we add $M_R$ (Memorization Ratio)~\cite{longpre-etal-2021-entity} to measure context faithfulness. To ensure reproducibility, all models were evaluated using deterministic decoding (temperature $=$ 0). Further metrics and details are in the Appendix~\ref{subsec:appendix-evaluation}.


\begin{table*}[t]
\centering
\caption{The experimental results for non-knowledge-conflict scenarios on LLaMA 3.1-8B-Instruct. The numbers in parentheses (e.g., -3.5) indicates the accuracy change compared to the full method and the best result is \underline{underlined}}
\label{tab:main-experient-2}
\resizebox{1.\linewidth}{!}{
\begin{tabular}{cccccccc}
\hline
 & \multicolumn{7}{c}{Methods} \\ \cline{2-8} 
 &  & \multicolumn{3}{c}{Context-faithful Prompting} & \multicolumn{2}{c}{Context-faithful Decoding} &  \\ \cline{3-7}
\multirow{-2}{*}{Dataset} & \multirow{-2}{*}{Origin} & Opin (Instr) & ATTR & KRE & CAD & COIECD & \multirow{-2}{*}{FaithfulRAG (Ours)} \\ \hline
MuSiQue-golden & 84 & 83(-1) & 80.4(-3.6)& 38.3$^*$ & 81.9(-3.1) & 83.3(-0.7) & {\color[HTML]{333333} {\ul \textbf{85.3}} \textbf{(+1.3)}} \\ \hline
SQuAD-golden & 95.2 & 96(+0.8) & 94.2(-1)& 81.4(-13.8)& 91.1(-4.1) & 95.1(-0.1) & {\ul \textbf{96.6}} \textbf{(+1.4)} \\ \hline
\end{tabular}
} \\
\footnotesize{$^*$ There is a sharp decline in performance as the model refuses to generate responses. }
\vspace{-3mm}
\end{table*}

\subsection{Main Results (Q1)}
To address \textbf{Q1}, we evaluate FaithfulRAG by comparing it to state-of-the-art baselines on four benchmark datasets, with main results shown in Tables~\ref{tab:main-experient} and ~\ref{tab:main-experient-2}, while MR results in the Appendix~\ref{appendix:mr-exp}. We summarize the observations as follows.

\noindent\textbf{Obs. 1.} FaithfulRAG consistently outperforms baseline models on four benchmark datasets. On FaithEval, FaithfulRAG with the Mistral-7B backbone achieves the highest score (81.7\%), surpassing the strongest baseline (73.2\% from KRE) by 8.5\%. Similarly, on SQuAD, FaithfulRAG (Llama3.1-8B) scores 86.3\%, exceeding the closest competitor (ChatQA-2.0 at 77.0\%) by 9.3\%. Besides, FaithfulRAG dominates MuSiQue and RealtimeQA with scores of 79.9\% and 84.1\%, demonstrating its robustness across diverse scenarios.

\noindent\textbf{Obs. 2.} FaithfulRAG demonstrates consistent and robust performance across diverse backbone LLMs. When evaluated on Llama3.1-8B, Qwen2.5-7B, and Mistral-7B, FaithfulRAG achieves state-of-the-art results while maintaining minimal performance variance, unlike methods such as KRE or CAD, which exhibit severe instability across different LLMs. This backbone-agnostic efficacy highlights its ability to harmonize parametric and contextual knowledge dynamically, regardless of model architecture and context complexity.

\noindent\textbf{Obs. 3.} FaithfulRAG achieves consistent performance in both knowledge conflict and non-conflict scenarios. As shown in Table~\ref{tab:main-experient-2}, FaithfulRAG also achieves the highest accuracy in non-conflict scenarios compared to all competitors. On MuSiQue-golden, FaithfulRAG achieves 85.3\%, outperforming strongest competitors like ATTR (80.4\%) and CAD (81.9\%). On SQuAD-golden, it scores 96.6\%, surpassing KRE (81.4\%) and COIECD (95.1\%). This aligns with the requirement for faithful systems to avoid performance degradation in non-conflict scenario~\cite{coiced}.


\begin{figure*}[t]
  \centering
  \includegraphics[width=1\textwidth]{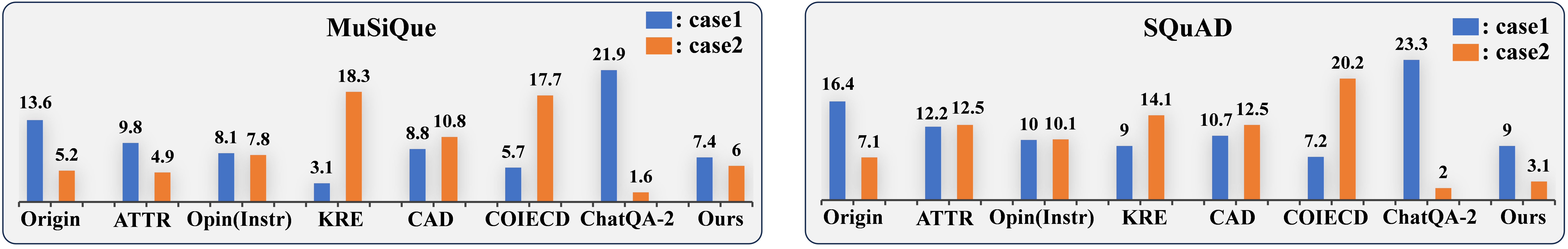} 
  \vspace{-5mm}
  \caption{The error distribution on MuSiQue and SQuAD datasets with Llama3.1-8b-instruct as the backbone LLM.
  }
  \label{fig:analyze2}
  \vspace{-3mm}
\end{figure*}

\begin{table*}[t]
\centering
\caption{Ablation study.
Numbers in parentheses (e.g., -1.9) represent the change in accuracy relative to full model.}
\vspace{-3mm}
\label{tab:ablation}
\resizebox{1.\linewidth}{!}{
\begin{tabular}{ccccccc}
\hline
\multirow{2}{*}{Module/Aspect} & \multirow{2}{*}{Variant} & \multicolumn{4}{c}{Dataset} & \multirow{2}{*}{Average}  \\ \cmidrule{3-6} 
\multicolumn{2}{c}{} & Faitheval & RealtimeQA & MuSiQue & SQuAD &  \\ \cmidrule(lr){1-7}
\multirow{2}{*}{ \makecell{Knowledge \\ Externalization} } & w/o Self-Context Generation & 77.2 & 77.9 & 79.9 & 85.1 & 80.0 (-1.90) \\
 & w/o Self-Knowledge Extraction & 77.9 & 80.6 & 79.3 & 85.2 & 80.8 (-1.10) \\ \cmidrule(lr){1-7}
\multirow{3}{*}{\makecell{Self-Think} } & w/o whole Module & 50.3 & 67.2 & 63.7 & 57.8 & 59.8 (-22.2) \\
 & w/o Think & 79.7 & 69.0 & 73.7 & 78.5 & 75.2 (-6.70) \\
 & w/o Reasoning & 79.6 & 73.5 & 72.2 & 78.7 & 76.0 (-5.90) \\
 \cmidrule(lr){1-7}
\multirow{2}{*}{CoT Influence} & Only CoT & 70.2 & 64.6 & 52.7 & 71.8 & 64.8 (-17.1) \\
 & w/o CoT & 82.1 & 79.6 & 78.7 & 78.9 & 79.8 (-2.10) \\ \cmidrule(lr){1-7}
 Full model & - & 79.8 & 81.4 & 79.9 & 86.3 & 81.9 \\ \bottomrule
\end{tabular}}
\vspace{-3mm}
\end{table*}

\subsection{Error Analysis (Q2)}

As discussed in Section~\ref{content:prior}, current faithful methods enhance faithfulness at the expense of an increased risk of context misinterpretation. In this section, we systematically analyze how effective  FaithfulRAG is in alleviating different types of errors. As shown in Figure~\ref{fig:analyze}, we have the following observations.

\noindent\textbf{Obs. 4.} Existing faithful models achieve desirable performance in alleviating Case 1 errors but at the cost of amplifying Case 2 errors. Specifically, COIECD achieved the best performance for Case 1 optimization, reducing Case 1 errors by an average of 8.6\%. However, this improvement came at the cost of a sharp increase in Case 2 errors, with the highest observed rise reaching 12.8\%. While Opin( Instr) demonstrated a more balanced performance, reducing Case 1 errors by an average of 6.0\%, while Case 2 errors only increased by 2.8\%. This trade-off stems from their inability to dynamically reconcile discrepancies between parametric and contextual knowledge.

\noindent\textbf{Obs. 5.} FaithfulRAG achieves balanced mitigation of both Case 1 and Case 2 errors, reducing them by 6.8\% and 1.6\%, respectively. This improvement stems from our well-designed framework, which enables LLMs to dynamically reconcile parametric knowledge with contextual evidence. By isolating discrepancies at the fact level and applying a self-think module, FaithfulRAG preserves high-quality parametric knowledge while systematically rejecting contexts that introduce logical inconsistencies or semantic divergence. 
\subsection{Alation Study (Q3)}
To evaluate the contributions of FaithfulRAG’s core components, we systematically ablate three key components: (i) Self-Fact Mining, (ii) Self-Think module, and (iii) Chain-of-Thought (CoT), generating \textbf{7} variants as shown in Table~\ref{tab:ablation}. Each variant was tested under knowledge-conflict scenarios to isolate its impact on performance. We have the following findings.



\noindent\textbf{Obs. 6.} The ablation of Self-Knowledge Extraction and Self-Context Generation reveals that both modules are critical for precise knowledge alignment. Removing Self-Knowledge Extraction degrades the model’s ability to analyze questions comprehensively, leading to a 1.1\% average accuracy drop, as the LLM fails to identify relevant parametric facts. Conversely, removing Self-Context Generation—which converts abstract self-knowledge into actionable context—causes a larger 1.9\% accuracy decline, demonstrating that raw parametric claims lack utility without contextual grounding. 

\noindent\textbf{Obs. 7.}  Ablating the full Self-Think module results in a 22.2\% average accuracy drop, as simply prepending self-aligned context to the original context fails to resolve conflicts dynamically. Replacing Think stage with Special Annotation (explicitly marking key facts) reduces accuracy by 6.7\%, proving that passive highlighting cannot replicate active reasoning. Similarly, substituting structured reasoning with naive Chain-of-Thought (CoT) decreases accuracy by 5.9\%, confirming that explicit guidance (e.g., conflict anticipation steps) is essential. These results validate that Self-Think’s diagnostic workflow (not just attention mechanisms or generic reasoning) drives reliable conflict resolution.



\noindent\textbf{Obs. 8.} While CoT enhances reasoning in some domains, naive application under knowledge conflicts causes a 17.1\% accuracy drop, as the LLM grows distrustful of conflicting contexts. FaithfulRAG addresses this by integrating CoT with conflict anticipation: the LLM first identifies potential discrepancies in self-aligned contexts before reasoning, mitigating distrust. Notably, even without CoT, FaithfulRAG’s accuracy decreases by only 2.1\%, demonstrating its robustness. This highlights that CoT must be tailored to handle knowledge conflicts, and FaithfulRAG’s structured reasoning framework achieves this adaptation.

To summarize, the combined ablation results reveal that FaithfulRAG’s components operate synergistically. Self-Fact Mining provides the foundation for precise fact alignment, while Self-Think enables dynamic conflict resolution. Without either, the model reverts to the limitations of suppression-based methods (e.g., over-reliance on context or parametric knowledge). This interdependence validates the framework’s design hypothesis: diagnostic reconciliation of conflicts instead of suppression is key to balancing faithfulness and accuracy.

\section{Related Work}
\textbf{Knowledge Conflict.} 
When encountering an external context with conflicting knowledge, an LLM tends to ignore such context~\cite{bi2024context,longpre-etal-2021-entity,bi2024factuality,jiang2025anyedit,fang2024alphaedit}.
A study indicates that the greater the divergence between retrieved information and the model's prior knowledge, the more likely the model is to ignore the retrieved information.~\cite{wu2024faithful}.
Another study~\cite{xie2024adaptive} further points out that when provided with both supporting and opposing evidence against their parametric memory, LLMs exhibit strong confirmation bias, tending to adhere to their parametric knowledge. Additionally, the study~\cite{ming2025faitheval} proposes the FaithEval framework to assess the faithfulness of LLMs across different contextual scenarios, revealing that even state-of-the-art models still struggle with the counterfactual context.

\noindent\textbf{Context-Faithful Method.}
Some methods enhance context faithfulness through fine-tuning. KAFT~\cite{li-etal-2023-large} fine-tune small parameter models like T5 using counterfactual contexts, SFR-RAG ~\cite{nguyen2024sfr}is a 9B models trained with various RAG-domain datasets. However, such fine-tuning approaches require substantial computational resources and are often difficult to transfer to other models. Among non-fine-tuning approaches, one category focuses on carefully designed prompt templates. For example, One method reconstruct the context as the narrator's statement and then inquire about the narrator’s viewpoint, encouraging the model to seek external perspectives~\cite{zhou-etal-2023-context}. Another method employs "Role Play" interventions to alter LLMs' decision-making styles when facing knowledge conflicts.~\cite{ying-etal-2024-intuitive} However, these methods heavily rely on the model’s inherent reasoning abilities and often suffer from limited generalizability. Another category of non-fine-tuning approaches modifies the model’s decoding strategy to improve its reliance on context. CAD\cite{cad}contrasts output probabilities with and without context to enhance contrastive decoding, amplifying the contextual probability distribution without accounting for conflicting contexts. In contrast, COIECD\cite{coiced} detects knowledge conflicts by measuring entropy changes during generation and dynamically adjusts the decoding strategy accordingly.

\section{Conclusions}
Large language models (LLMs) often generate hallucinations (factually inconsistent or unfaithful contents).  While retrieval-augmented generation (RAG) has shown promise in enhancing language models' capabilities through external knowledge integration, maintaining faithfulness to retrieved contexts remains a significant challenge. 
This paper identifies and analyzes critical unfaithfulness issues that emerge during RAG's generation phase, particularly when the model's parametric knowledge contradicts retrieved information. 
Existing faithful RAG methods enforce alignment with the retrieved context by suppressing the model's parametric knowledge, but often increase the risk of misinterpreting the context.
To address these challenges, we propose a novel faithful framework (FaithfulRAG) that explicitly identifies and resolves knowledge conflicts at the fact level. FaithfulRAG first concretizes the model's parametric knowledge at the fact level and then identifies conflicting knowledge by aligning self-fact with contexts. After that, FaithfulRAG designs a self-think module, allowing LLMs to reason about and integrate conflicting facts before generation. Experiments show that FaithfulRAG outperforms the strongest competitors, generating accurate and faithful responses.
\section*{Limitations}
While FaithfulRAG advances error mitigation in text-based retrieval-augmented generation, its current scope is limited to textual inputs and does not yet support multimodal information (e.g., images, audio, or structured data). Extending the framework to incorporate multimodal input would enable a more comprehensive assessment of how models navigate and resolve conflicts between heterogeneous knowledge sources. Given that real-world information is often multimodal, such an extension could improve FaithfulRAG’s ability to detect and reconcile inconsistencies that arise from modality-specific biases or divergent contextual signals.

In addition, multimodal integration would enhance the applicability of the framework to domains where information synthesis from multiple sources is crucial, such as medical diagnosis (where textual reports, imaging data and patient history must be jointly considered), autonomous systems (which rely on the integration of visual, auditory, and textual signals for decision making), and interactive AI (where understanding user intent often involves processing speech, gestures, and textual input). Addressing these challenges would not only improve FaithfulRAG’s robustness but also contribute to broader advancements in multimodal retrieval-augmented generation, cross-modal reasoning, and trustworthy AI.

\section*{Ethics Statement}
We confirm that we have fully complied with the ACL Ethics Policy in this study. Our research utilizes four publicly available datasets:FaithEval, MuSiQue, SQuAD, and RealtimeQA.FaithEval is designed to evaluate LLM and RAG faithfulness across Unanswerable, Inconsistent, and Counterfactual contexts. MuSiQue and SQuAD, sourced from KRE, assess LLM robustness against knowledge conflicts in modified MRC and commonsense reasoning tasks. RealtimeQA is a dynamic QA dataset for evaluating models' ability to handle real-time information. All datasets used in this study have been extensively employed in retrieval-augmented generation research and do not contain private, sensitive, or personally identifiable information. We carefully select these datasets to ensure ethical compliance and to mitigate potential biases. Our study does not involve the collection or modification of user-generated content, nor does it introduce synthetic data that could lead to unintended misinformation.
\section*{Acknowledgements}
The project was supported by 
National Key R\&D Program of China (No. 2022ZD0160501), 
Natural Science Foundation of Fujian Province of China (No. 2024J011001),
and
the Public Technology Service Platform Project of Xiamen (No.3502Z20231043).
We also thank the reviewers for their insightful comments.

\bibliography{custom}

\clearpage
\appendix

\begin{table*}[ht]
\small
\centering
\renewcommand{\arraystretch}{1.3} 
\setlength{\extrarowheight}{1pt}  
\begin{tabularx}{\textwidth}{|p{2.2cm}|X|}
    \hline
    \textbf{Query} & An astronomer observes that a planet rotates faster after a meteorite impact. Which is the most likely effect of this increase in rotation? \\ 
    \hline
    \textbf{Context} & Observing planetary changes post-meteorite impacts has always been a critical aspect of studying celestial mechanics. When a substantial meteorite strikes a planet, it can significantly alter the planet's rotation speed. This phenomenon was notably observed with the planet Tranus-4 in a neighboring galaxy…
 \\ 
    \hline
    \textbf{Self-Knowledge} \newline \textbf{Extraction} & \textbf{Planetary rotation and angular momentum:} Understanding how a planet's rotation is affected by its angular momentum, which is a measure of an object's tendency to continue rotating. This knowledge is necessary to explain how the meteorite impact could cause the planet to rotate faster.
    
    \textbf{Conservation of angular momentum:} Recognizing that the conservation of angular momentum is a fundamental principle in physics, which states that the total angular momentum of a closed system remains constant over time. This principle is crucial in understanding how the planet's rotation speed changes after the meteorite impact
    
    \textbf{...}
    \\ 
    \hline
    \textbf{Self-Context} \newline \textbf{Generation} & A meteorite impact can transfer momentum to a planet, potentially increasing its rotation speed. This effect is due to the conservation of angular momentum, a fundamental principle in physics that states the total angular momentum of a closed system remains constant over time. As the meteorite impacts the planet, its momentum is transferred, causing the planet's rotation to accelerate... \\ 
    \hline
    \textbf{Self-Fact} \newline \textbf{Extraction} & 1.A meteorite impact can transfer momentum to a planet, potentially increasing its rotation speed.\newline 2.The conservation of angular momentum is a fundamental principle in physics that states the total angular momentum of a closed system remains constant over time.\newline 3.The momentum of a meteorite is transferred to a planet upon impact, causing its rotation to accelerate.\newline \textbf{...}\\ 
    \hline
    \textbf{Contextual} \newline \textbf{Alignment} & ...Normally, an increase in rotational speed would be closely associated with geophysical implications such as changes in atmospheric dynamics and magnetic field intensities. \textbf{Moreover, the sudden acceleration can cause a redistribution of mass within the planet, thus potentially affecting its gravitational pull.}Researchers from the Galactic Federation Science Division have recently published a detailed analysis demonstrating a correlation between rotation speed and gravitational strength...  \\ 
    \hline
    \textbf{Self-Think}& 
    1. [Fact Analysis]: Facts explicitly state They observed that as Tranus-4's rotation speed increased, the equatorial bulge became
 more pronounced, effectively increasing the planet's gravitational pull at the equator. \newline
    2. [Option Matching]: Option D directly matches the factual declaration \newline
    3. [Context Check]: No contextual supplementation needed - Facts provide conclusive evidence \newline
    4. [Final Verification]: No conflicting information.
    \\ 
    \hline
    \textbf{Output} & 
     Answer: Planetary gravity will become stronger. \\
     \hline
\end{tabularx}
\caption{A complete pipeline of our method on FaithEval using Llama3.1-8B-Instruct as the backbone model.}
\label{tab:appendix-example}
\end{table*}





\section{Frequently Asked Questions (FAQs)}
\subsection{Code and Dataset Availability}
To promote transparency and reproducibility, we have uploaded our code into the Anonymous GitHub at \textcolor{blue}{\url{https://github.com/DeepLearnXMU/Faithful-RAG.}}. The repository includes the source code of FaithfulRAG and scripts for data preprocessing, model training, and evaluation. Additionally, all datasets used in our experiments are also provided or linked within the repository, ensuring that researchers have full access to the resources required to reproduce and extend our work. 

\subsection{What are advantages of FaithfulRAG?}
FaithfulRAG introduces several key advancements over existing RAG systems and tailored designed faithful methods, addressing fundamental limitations while maintaining practical applicability:

\noindent\textbf{Superior performance on knowledge conflict scenarios.} Unlike existing faithful methods that reduce unfaithful errors (rigid adherence to parametric knowledge) at the cost of increasing incorrect match errors (blind acceptance of flawed contexts), FaithfulRAG resolves both error types simultaneously. Specifically, it reduces Case 1 errors by 6.8\% while simultaneously decreasing Case 2 errors by 1.6\% on average as shown in Figure~\ref{fig:analyze2}. This improvement arises from its fact-level conflict resolution and self-think mechanisms that enable LLMs to preserve high-quality parametric knowledge and reject semantically divergent contexts. 

\noindent\textbf{Robustness in Non-Conflict Scenarios.} A good faithful model should improve performance in knowledge-conflict scenarios while maintaining desirable performance in non-conflict scenarios.  
While most existing methods suffer in non-conflict settings (e.g., KRE’s 13.8\% drop on SQuAD-golden), our FaithfulRAG achieves the best performance on both SQuAD-golden and MuSiQue-golden. It gets 96.6\% on SQuAD-golden (+1.4\% over origin) and 85.3 on MuSiQue-golden (+1.3\%), proving its robustness and capability to avoid overfitting to either LLM's parametric knowledge or incorrect contextual information.

\noindent\textbf{Backbone-Agnostic Consistency.} FaithfulRAG achieves stable performance across diverse backbone LLMs (e.g., Llama3.1-8B, Qwen2.5-7B, Mistral-7B), with minimal variance in accuracy. For instance, on the SQuAD dataset, it attains 86.3\% accuracy with Llama3.1-8B and 85.7\% with Mistral-7B, outperforming task-specific models like ChatQA-2.0 by 9.3\% as shown in Table~\ref{tab:main-experient}. This consistency ensures reliable deployment across heterogeneous environments in practice.
\begin{algorithm}[t]
    \caption{ The Workflow of FaithfulRAG.}
    \label{alg:algorithm}
    \textbf{Input}: Question \(Q\), Original context \(\mathcal{C}_{\text{orig}}\).\\
    \textbf{Output}: Answer \(a\).\\
    \begin{algorithmic}[1] 
        \STATE \textbf{Self-Fact Mining} \\
        \STATE Extract a set of high-level knowledge from \(Q\); \\
         $\mathcal{K}_{\text{self}}(Q) = \{k_1, k_2, \dots, k_n\}$ 
        \STATE Generate context \(\mathcal{C}_{\text{self}}(Q)\) grounded in \(\mathcal{K}_{\text{self}}(Q)\); \\
        $\mathcal{C}_{\text{self}}(Q)=\mathcal{G}_{1}(Q,\mathcal{K}_{\text{self}}(Q))$
        \STATE Extract facts \(\mathcal{F}_{\text{self}}(Q)\) from \(\mathcal{C}_{\text{self}}(Q)\); \\
        $\mathcal{F}_{\text{self}}(\mathcal{C}_{\text{self}}) = \{f_1, f_2, \dots, f_m\}$
        \STATE \textbf{Contextual Knowledge Alignment} \\
        \STATE Divide the original context \(\mathcal{C}_{\text{orig}}\) into fixed-size chunks \(\mathcal{C}_{\text{orig}}^{j}\);\\
        $\mathcal{C}_{\mathrm{orig}} = \bigcup_{i=1}^{m} \mathcal{C}_{\text{orig}}^{i}.$
        \FOR{each fact \(f\) in \(\mathcal{F}_{\text{self}}(Q)\)}
            \FOR{each context chunk \(c\) in \(\mathcal{C}_{\text{orig}}\)}
                \STATE Compute similarity between \(f\) and \(c\) \\
                $Sim(\textbf{f}, \textbf{c}) = \text{cos}(\textbf{f}, \textbf{c}).$
            \ENDFOR
        \ENDFOR
        \STATE Select top-\(k\) chunks as \(\mathcal{C}_{\text{aligned}}\)
        \STATE \textbf{Self-Think} \\
        \IF{\(\mathcal{C}_{\text{aligned}}\) is reliable}
            \STATE Generate \(a\) based on \(\mathcal{C}_{\text{aligned}}\)
        \ELSE
            \STATE Generate \(a\) using \(\mathcal{R}_{\text{STR}}\bigl(\mathcal{C}_{\text{aligned}}, \mathcal{C}_{\text{orig}}\bigr)\)
        \ENDIF
        \STATE \textbf{return} \(a\).
    \end{algorithmic}
\end{algorithm}

\subsection{Why focusing on the generation stage of RAG, and why is it crucial to maintain the faithfulness of outputs?}
Despite recent advances in retrieval mechanisms, a persistent challenge remains: a significant gap exists between the quality of the retrieved context and the factual accuracy of the generated output. While the context retrieved by the model may be highly relevant, the generative models often struggle to faithfully incorporate this context, resulting in generated text that may diverge from factual correctness. Recent empirical evidence reveals the critical gap between retrieval quality and generation faithfulness. 
The generation stage determines how this context is interpreted and integrated with the model’s parametric knowledge. Existing methods often fail at this stage due to (i) Over-confidence: many RAG systems rigidly prioritize the retrieved context by suppressing parametric knowledge, leading to blind acceptance of flawed or conflicting information. (ii) Logical incoherence: Without critical reasoning, models struggle to resolve contradictions, resulting in inconsistent or unsafe outputs. By targeting generation, in this paper, we address the root cause of unfaithfulness and guide the LLM to dynamically reconcile knowledge conflicts.

\begin{table*}[htp]
\centering
\caption{The experiment results of error analysis on LLaMA 3.1-8B-Instruct. The best result is \underline{underlined}.}
\label{tab:analysis_result}
\resizebox{1.0\linewidth}{!}{
\begin{tabular}{cccccccccc}
\hline
\multicolumn{2}{c}{\multirow{3}{*}{Method}} & \multicolumn{4}{c}{MuSiQue} & \multicolumn{4}{c}{SQuAD} \\ \cline{3-10} 
\multicolumn{2}{c}{} & Case 1 & Case 2 & Case 3 & All & Case 1 & Case 2 & Case 3 & All \\ \cline{3-10} 
\multicolumn{2}{c}{} & \multicolumn{1}{l}{Error rate} & \multicolumn{1}{l}{Error rate} & \multicolumn{1}{l}{Error rate} & \multicolumn{1}{l}{Error rate} & \multicolumn{1}{l}{Error rate} & \multicolumn{1}{l}{Error rate} & \multicolumn{1}{l}{Error rate} & \multicolumn{1}{l}{Error rate} \\ \cline{1-10} 
Origin & - & 13.6 & 5.2 & 13.3 & 32.1 & 16.4 & 7.1 & 7 & 30.5 \\ \hline
\multirow{3}{*}{Prompting} & Opin (instr) & 8.1 & 7.8 & 13.7 & 29.6 & 12.2 & 12.5 & 5.5 & 26.6 \\ \cline{2-10} 
 & ATTR & 9.8 & 4.9 & 22.4 & 37.1 & 10 & 10.1 & 6.5 & 30.2 \\ \cline{2-10}
 & KRE & 3.1 & 18.3 & 49.3$^*$ & 70.7$^*$ & 9 & 14.1 & 14.9 & 38 \\
 \cline{2-10} 
 & \textbf{Average} & \textbf{9.0} & \textbf{6.4} & \textbf{18.1} & - & \textbf{11.1} & \textbf{11.3} & \textbf{6.0} & - \\ \hline
\multirow{3}{*}{Decoding} & CAD & 8.8 & 10.8 & 7.8 & 27.4 & 10.7 & 12.5 & 5.5 & 28.7 \\ \cline{2-10} 
 & COIECD & 5.7 & 17.7 & 10.7 & 34.1 & 7.2 & 20.2 & 7.6 & 35.0 \\ \cline{2-10} 
 & \textbf{Average} & {\ul \textbf{7.3}} & \textbf{14.3} & \textbf{9.3} & - & {\ul \textbf{9}} & \textbf{16.4} & \textbf{6.6} & - \\ \hline
 RAGs & ChatQA-2.0 & 21.9 & {\ul 1.6} & 9.8 & 33.3 & 23.3 & {\ul 2} & 9.7 & 35.0 \\ \hline
Ours & FaithfulRAG & \textbf{7.4} & \textbf{6} & {\ul \textbf{8}} & {\ul \textbf{21.4}} & {\ul \textbf{9}} & \textbf{3.1} & {\ul \textbf{3.4}} & {\ul \textbf{15.5}} \\ \hline
\end{tabular}
}    
\end{table*}
\begin{figure*}[htp]
  \centering
  \includegraphics[width=1\textwidth]{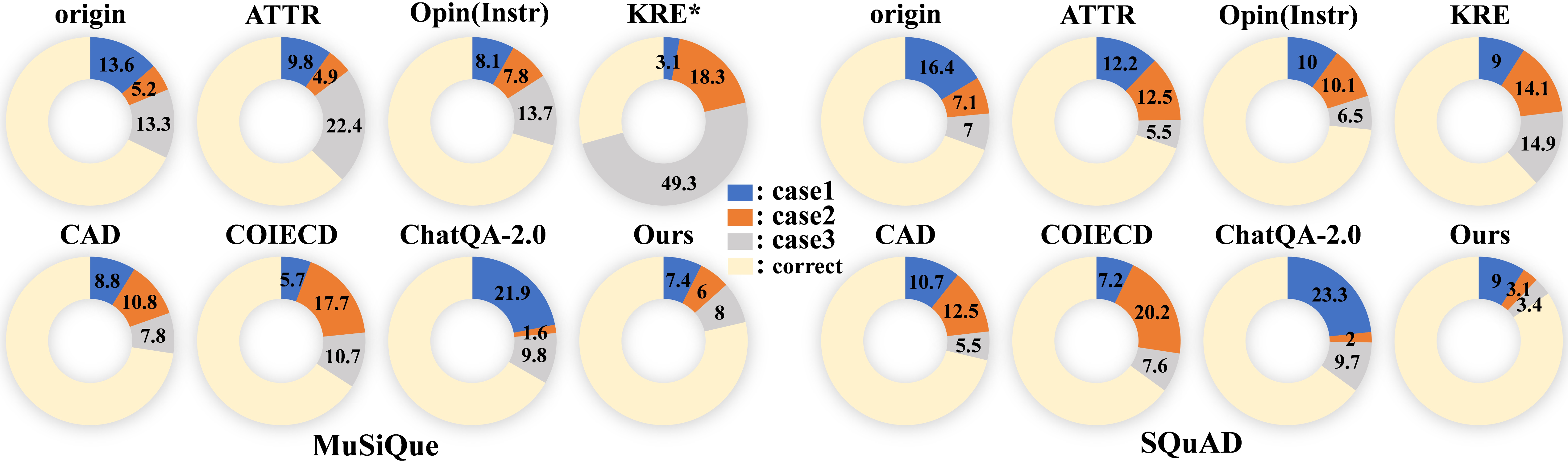} 
  \caption{The error distribution on MuSiQue and SQuAD datasets with Llama3.1-8b-instruct as the backbone LLM.
  Case 1 and Case 2 are consistent with Section~\ref{content:prior}, while Case 3 represents all other scenarios beyond these two cases.
  }
  \label{fig:analyze}
\end{figure*}

\subsection{How do we get the datasets in scenarios with and without knowledge conflicts?}\label{subsec:appendix-prior}

We use the MuSiQue and SQuAD datasets from the KRE~\cite{ying-etal-2024-intuitive} to evaluate LLM performance in both knowledge conflict and non-conflict scenarios. These datasets contain two types of contexts: (i) Negative context, where correct entities are replaced with incorrect ones, simulating a knowledge conflict scenario. (ii) Golden context, which remains unaltered and serves as an approximation of a non-knowledge conflict scenario. It is important to note that while the golden context may still contain some degree of knowledge conflict, its occurrence rate is significantly lower, allowing us to approximate it as a non-conflict dataset.
\subsection{How do we calculate the distribution of different error cases?}
To measure the error distribution, we follow a step-by-step approach to compute the frequency of different error cases. (i) Parametric prediction: We first generate a prediction from the LLM without providing any context, relying solely on its parametric knowledge. (ii) Contextual prediction: We then provide the corresponding context and generate a new prediction based on the context from the LLM. (iii) Case classification: We classify errors by comparing the two predictions:
\begin{itemize}
    \item Over-confidence Error (Case 1): If the parametric prediction and contextual prediction are identical, the model is considered to have ignored the retrieved context and relied on its internal parametric knowledge.
    \item Incorrect-match Error (Case 2): If the contextual prediction differs from the parametric prediction, and the prediction appears in the provided context but is incorrect, the model is considered to have been misled by the context.
\end{itemize}

\section{Additional Experiments}
\subsection{Case Study (Q4)}\label{appendix:case}
To empirically validate the contribution of each component in FaithfulRAG, we conduct a granular case study using a representative instance from the FaithEval dataset, with Llama3.1-8B-Instruct as the backbone model. The intermediate outputs at each stage are detailed in Table~\ref{tab:appendix-example}, illustrating how our framework progressively resolves knowledge conflicts while maintaining logical coherence.

\paragraph{Step 1: Self-Knowledge Extraction.} 
In this phase, the model extracts abstract parametric knowledge relevant to the query.

\paragraph{Step 2: Self-Context Generation.}  The model synthesizes its self-knowledge into a complete context that reflects the LLM's "self-perception" about the query-related facts. This bridges abstract knowledge with task-specific reasoning, enabling downstream conflict resolution.

\paragraph{Step 3: Self-Fact Extraction.}
The model distills self-context into fine-grained self-facts. These self-facts serve as anchors for aligning with the retrieved context while preserving logical constraints.

\paragraph{Step 4: Contextual Alignment}
The model first converts self-facts and segmented retrieved context into embeddings, then computes the similarity between these embeddings. Finally, it selects the top-k chunks to form a self-aligned context that closely aligns with self-facts.

\paragraph{Step 5: Self-Think.}
Finally, the model reconciles self-aligned context with the whole context through iterative reasoning, including (i) Flag the contradiction between guidelines and the prescription. (ii) Probes for missing data in the context. (iii) Verify specialist approval and monitor renal function.
\begin{table}[t]
\centering
\caption{The MR metric results of our method and the SOTA baseline across four datasets. Lower MR values indicate higher context faithfulness.
}
\vspace{-3mm}
\resizebox{1.0\linewidth}{!}{
\begin{tabular}{cc|cc|cc} 
\midrule
\multirow{2}{*}{Model} & \multirow{2}{*}{Backbone LLM} & \multicolumn{4}{c}{Dataset} \\ 
\cmidrule(lr){3-6} 
& & \multicolumn{2}{c|}{MR} & \multicolumn{2}{c}{ACC} \\ 
& & MuSiQue & SQuAD & MuSiQue & SQuAD \\ 
\midrule
 \multicolumn{6}{c}{\textbf{Group 1: With Full Context}} \\ \cmidrule(lr){1-6}
\multirow{2}[6]{*}{Origin} &  llama3.1-8b-instruct &17.6 & 20.7 & 67.8 & 69.5\\
 & qwen2.5-7b-instruct  &17.3 & 27.6 & 75.2 & 68.3\\ 
 & mistral-7b-instruct  &15.9& 25 & 67.6 & 67.2\\
\cmidrule(lr){1-6}

\multicolumn{6}{c}{\textbf{Group 2: Specific RAG Models}} \\ \cmidrule(lr){1-6}
Self-RAG& Llama2-7B & 18.8 &30.1 & 54.1 & 62.0\\
ChatQA-1.5& Llama3.1-8B & 17.7&  19.1 &75.0 &77.0\\
ChatQA-2.0~ & Llama3.1-8B & 15.5 &20.8 &77.2&75.4\\
\cmidrule(lr){1-6}

\multicolumn{6}{c}{\textbf{Group 3: Context-faithful Prompting}} \\ \cmidrule(lr){1-6}
\multirow{2}[6]{*}{Opin(Instr)} &llama3.1-8b-instruct & 11.3 &13.8 &70.3 &73.4\\
&qwen2.5-7b-instruct & 15.3 & 25.5 &76.9 &70.5\\
&mistral-7b-instruct & 13.1 & 22.9 &68.1 &69.3\\
\cmidrule(lr){2-6}
\multirow{2}[6]{*}{ATTR} &llama3.1-8b-instruct & 14.7 & 16.8& 62.8 &69.5\\
&qwen2.5-7b-instruct & 14 &22.2 &\underline{78.7}&72.9 \\
&mistral-7b-instruct & 12.7 &20.6 &66.1 &70.2\\
\cmidrule(lr){2-6}
\multirow{2}[6]{*}{KRE} &llama3.1-8b-instruct & 14.0 &14.5 &35.9$^*$ &66.1\\
&qwen2.5-7b-instruct & 18.6 & 22.7 &70.7 &73.7\\
&mistral-7b-instruct & 16.3 & 16.8 &50.6 &74.6\\

\cmidrule(lr){1-6}

\multicolumn{6}{c}{\textbf{Group 4: Context-faithful Decoding}} \\ \cmidrule(lr){1-6}
\multirow{2}[6]{*}{CAD} &llama3.1-8b-instruct & 12.8& 14.6 &72.6 & 71.2\\
&qwen2.5-7b-instruct & 12.4 & 21.7 & 78.6 &73.4\\
&mistral-7b-instruct & 11.8 & 18 & 63.6 &66.9 \\
\cmidrule(lr){2-6}
\multirow{2}[6]{*}{COIECD} &llama3.1-8b-instruct & \underline{9.2} & \underline{10.7} & 70.5 & 71.8\\
&qwen2.5-7b-instruct & \underline{9.8} & \underline{15.5} & 69.7 & 70.8 \\
&mistral-7b-instruct & 10.3 & 16.1 & 66.8 & 65.4\\

\cmidrule(lr){1-6}
\multirow{2}[6]{*}{Ours} &llama3.1-8b-instruct & 11.8 & 13.5 & \textbf{\underline{79.9}} & \textbf{\underline{86.3}}\\
&qwen2.5-7b-instruct & 13.8 & 15.7 & 78.0 & \underline{78.3} \\
&mistral-7b-instruct & \textbf{\underline{7.7}} & \textbf{\underline{9.7}} & \underline{78.5} & \underline{85.7}\\
\bottomrule
\end{tabular}
}
\label{appendix-mr}
\vspace{-5mm}
\end{table}

\subsection{Error Analysis}
Current faithful methods enhance faithfulness at the expense of an increased risk of context misinterpretation. In this section, we systematically analyze how effective FaithfulRAG is in alleviating different types of errors. Specifically, we consider \textbf{3} different types of errors, where Case 1 and Case 2 are over-confidence and incorrect-match errors, respectively, consistent with Section~\ref{content:prior}, and Case 3 represents all other scenarios beyond these two cases. As shown in Table~\ref{tab:analysis_result} and Figure~\ref{fig:analyze}, we have the following findings. FaithfulRAG achieves balanced mitigation of both Case 1 and Case 2 errors. Specifically, it reduces Case 1 errors by 6.8\% while simultaneously decreasing Case 2 errors by 1.6\% on average.
This improvement stems from our well-designed framework, which enables LLMs to dynamically reconcile parametric knowledge with contextual evidence. By isolating discrepancies at the fact level and applying a self-think module, FaithfulRAG preserves high-quality parametric knowledge while systematically rejecting contexts that introduce logical inconsistencies or semantic divergence. FaithfulRAG aslo demonstrates superior performance in Case 3, maintaining the lowest error rate (5.7\% ) compared to baselines, demonstrating its robustness in handling edge cases. 
\begin{table}[t]
\centering
\caption{Model Analysis. The comparison of performance between our model and SOTA baselines by integrating different sizes of LLMs. 
}
\vspace{-2mm}
\resizebox{1.0\linewidth}{!}{
\begin{tabular}{cc|cccc}
\midrule
 \multirow{2}{*}{Model} & \multirow{2}{*}{Backbone LLM} & \multicolumn{4}{c}{Dataset} \\ 
 \cmidrule(lr){3-6}
 & &FaithEval & RealtimeQA &MuSiQue & SQuAD \\ 
\midrule
\multicolumn{6}{c}{\textbf{Group 1: Without Context}} \\ \cmidrule(lr){1-6}
\multirow{2}[6]{*}{Origin} & llama3.2-3b-instruct & 15.3 & 41.6 & 7.6 & 7.7\\
& llama2-13b-instruct  &17.5 &15.9 &18.7 &14.2 \\ 
 & deepseek-moe-16b  &14.1 &8.9 &7.3 &10.7 \\ 
 \cmidrule(lr){1-6}
 \multicolumn{6}{c}{\textbf{Group 2: With Full Context}} \\ \cmidrule(lr){1-6}
\multirow{2}[6]{*}{Origin} &  llama3.2-3b-instruct &67.1 &80.5 &66.1 & 79.2\\
 & llama2-13b-instruct  &75.6 &55.7 &80.5 & 78.4 \\ 
 & deepseek-moe-16b  &60.3 &51.3 &67.4& 73.9 \\
\cmidrule(lr){1-6}


\multicolumn{6}{c}{\textbf{Group 3: Context-faithful Prompting}} \\ \cmidrule(lr){1-6}
\multirow{2}[6]{*}{Opin(Instr)} &llama3.2-3b-instruct & 65.7 &\textbf{\underline{84.9}}& 69.2 &75.2 \\
&llama2-13b-instruct & 67.9 &60.2& 81.5 & 79.1 \\
&deepseek-moe-16b & 62.8 & 60.2& 72 & 72 \\
\cmidrule(lr){2-6}
\multirow{2}[6]{*}{ATTR} &llama3.2-3b-instruct & \underline{70.6}  &77.9 &66.6 & 79.2\\
&llama2-13b-instruct & 76.2 &\underline{61.1} &81.5&78.9  \\
&deepseek-moe-16b & 59.2 &53.1 &67.1 &73.1 \\
\cmidrule(lr){1-6}


\multicolumn{6}{c}{\textbf{Group 4: Context-faithful Decoding}} \\ \cmidrule(lr){1-6}
\multirow{2}[6]{*}{CAD} &llama3.2-3b-instruct & 60.9 &65.5 &76.3& 76.8 \\
&llama2-13b-instruct & 78.1 &51.3 &80.4 & 75.6\\
&deepseek-moe-16b & 60.2 &49.6 &52.8 & 72.9  \\
\cmidrule(lr){2-6}
\multirow{2}[6]{*}{COIECD} &llama3.2-3b-instruct & 59.5 &63.7 &70.6 & 66.4\\
&llama2-13b-instruct & 78.1 &50.4 &80.0 & 75.4 \\
&deepseek-moe-16b & 68.9 & 51.3 & 69.2 & 76.0  \\

\cmidrule(lr){1-6}
\multirow{2}[6]{*}{ Ours} &llama3.2-3b-instruct & 70.1 &79.6 &\underline{78.4} & \underline{79.8}\\
&llama2-13b-instruct & \textbf{\underline{80.0}} &53.1 &\textbf{\underline{83.9}} &\textbf{\underline{86.3}} \\
&deepseek-moe-16b & \underline{79.2} &\underline{61.9} &\underline{76.1}&\underline{82.1} \\
\bottomrule
\end{tabular}
}\\
\vspace{-2mm}
\label{tab:appendix-experient}
\end{table}

\subsection{$M_R$ Result}\label{appendix:mr-exp}
Following previous work~\cite{longpre-etal-2021-entity}, we calculate the context faithfulness metric ($M_R$) on entity-level knowledge conflict datasets, MuSiQue and SQuAD. The result is shown in Table~\ref{appendix-mr}. Among all methods, our approach achieves the best performance on Mistral, with 7.7\% on MuSiQue and 9.7\% on SQuAD. COIECD performs best across different backbone models, as its modified decoding strategy enforces stronger context alignment, leading to higher context faithfulness. However, our method outperforms COIECD by up to 10\% in ACC, demonstrating that our approach not only improves context faithfulness but also maintains strong downstream task performance (ACC).

\subsection{Model Analysis}
To comprehensively demonstrate the generalizability of our method, we conduct experiments on models of different sizes and architectures with the results shown in Table~\ref{tab:appendix-experient}. Our method significantly improves performance across models of different scales, from smaller 3B models to larger 16B models. This demonstrates the strong generalizability and adaptability of our approach, making it effective across a wide range of LLM architectures. Note that the KRE method shows a high rate of refusal responses. Therefore, we do not include it as a baseline in our additional experiments.

\subsection{Impact of Different Embedding Models}
In the main experimental setup, we employ all-MiniLM-L6-v2\footnote{https://huggingface.co/sentence-transformers/all-MiniLM-L6-v2} to embed self-facts. To further assess the effect of different embedding models, we conducted additional experiments by replacing it with alternative models. As shown in Table \ref{appendix-embedding}, all-MiniLM-L6-v2 achieves the best performance on FaithEval and SQuAD, while TinyBERT-L6-v2 performs optimally on RealtimeQA and MuSiQue. We selected all-MiniLM-L6-v2 as the default embedding model in our main setup due to its smaller parameter size and balanced overall performance across benchmarks.

\begin{table}[t]
\centering
\caption{Performance Comparison of Different Embedding Models Using Llama3.1-8B-Instruct as the Backbone Model.}
\vspace{-2mm}
\resizebox{1.0\linewidth}{!}{
\begin{tabular}{cc|cccc}
\toprule
\textbf{Embedding Model} & \textbf{Params} & \textbf{Faitheval} & \textbf{RealtimeQA} & \textbf{MuSiQue} & \textbf{SQuAD} \\
\midrule
all-MiniLM-L6-v2 & 22.7M & 79.8 & 81.4 & 79.9 & 86.3 \\
TinyBERT-L6-v2 & 67M & 78.3 & 84.1 & 80.1 & 85.1 \\
\makecell[c]{contriever-\\sentencetransformer} & 110M & 76.9 & 83.2 & 79.9 & 86.3 \\
sentence-t5-base & 110M & 78.3 & 80.5 & 80.0 & 86.2 \\
\bottomrule
\end{tabular}
}
\vspace{-6mm}
\label{appendix-embedding}
\end{table}


\section{Implementation Details}
\subsection{Benchmark Dataset} \label{subsec:appendix-dataset}
We evaluate FaithfulRAG on four benchmark datasets, including FaithEval~\cite{ming2025faitheval}, RealtimeQA~\cite{kasai2024realtime}, and MuSiQue~\cite{trivedi2022MuSiQuemultihopquestionssinglehop},  SQuAD~\cite{rajpurkar2016SQuAD} from KRE(knowledge robustness evaluation)~\cite{ying-etal-2024-intuitive}.

\noindent\textbf{FaithEval~\cite{ming2025faitheval}:} A novel benchmark dataset designed to evaluate the faithfulness of LLM and RAG systems across various contextual scenarios. This dataset consists of 4,900 high-quality questions covering three task types: Unanswerable, Inconsistent, and Counterfactual contexts. The Counterfactual Context is constructed based on ARC-Challenge, a multiple-choice science QA dataset at the elementary school level. Its knowledge conflict extends beyond the entity level, involving more complex logical relationships. We selected the Counterfactual subset as it aligns closely with our motivation.

\noindent\textbf{RealtimeQA~\cite{kasai2024realtime}:} A dynamic question-answering dataset designed to evaluate the ability of QA systems to handle real-time information, challenging the assumptions of traditional static QA datasets and targeting more immediate application scenarios. To test the model performance in extreme cases where some contexts are irrelevant to the question, we follow the ~\cite{zhou-etal-2023-context} and construct the RealTime QA-22 dataset. This dataset selects six questions from the first week of 2022 as the test set, with an equal proportion of answerable and unanswerable questions.

\textbf{MuSiQue~\cite{trivedi2022MuSiQuemultihopquestionssinglehop} and  SQuAD~\cite{rajpurkar2016SQuAD}:} These two datasets are from the previous research~\cite{ying-etal-2024-intuitive} designed to study the behavior of LLMs when confronted with contexts that conflict with their internal memory. It encompasses tasks involving both factual knowledge and commonsense reasoning and is constructed by modifying existing machine reading comprehension and commonsense reasoning datasets. Each entry in the dataset contains a negative context with knowledge conflicts and an unmodified golden context. In the main experiment, we use the negative-context as the context to construct MuSiQue and SQuAD. While when evaluating performance in standard scenarios, the golden-context is used as the context.\footnote{Unless otherwise specified, MuSiQue and SQuAD refer to the corresponding datasets using negative context.} To better align with real-world RAG scenarios, we use the  subset with longer context lengths.

\subsection{Baseline Selection}\label{subsec:appendix-baselines}
We carefully select baselines from four categories for a comprehensive evaluation.

\noindent\textbf{Oringin Model:} This model serves as a standard baseline in existing RAG systems. It employs an instruction-following model, such as llama3.1-8b-instruct, as its base model. Given a question and its related context, the model generates accurate and contextually appropriate answers

\noindent\textbf{Specific RAG Systems:} This group of methods~\cite{asai2023selfraglearningretrievegenerate,xu2024chatqa,liu2024chatqa} provides solutions for various RAG application scenarios, aiming to enhance the general ability of RAG systems across multiple benchmarks. By carefully selecting instruction-following datasets and designing tailored Supervised Fine-tuning (SFT) strategies, they improve the performance of LLMs in retrieval-augmented tasks.

\noindent\textbf{Context-faithful Prompting:} This category of methods~\cite{zhou-etal-2023-context,ying-etal-2024-intuitive} enhance context faithfulness in LLMs by designing specialized prompting strategies. It addresses knowledge conflict issues by strengthening the reliance of LLMs on the provided context while reducing their dependence on parametric knowledge.

\newpage

\begin{figure*}[ht]
  \centering
  \begin{subfigure}[b]{1.\textwidth}
    \centering
    \includegraphics[width=\textwidth]{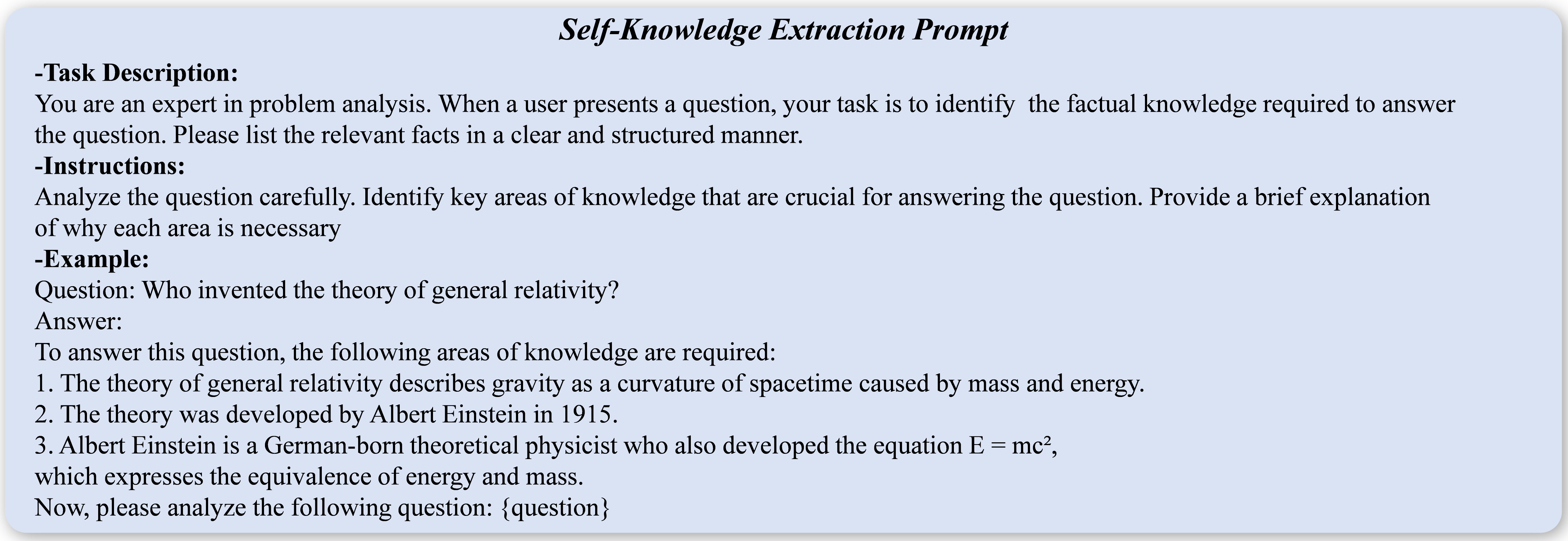}
    \label{fig:prompt1}
  \end{subfigure}
  
  
  \begin{subfigure}[b]{1.\textwidth}
    \centering
    \includegraphics[width=\textwidth]{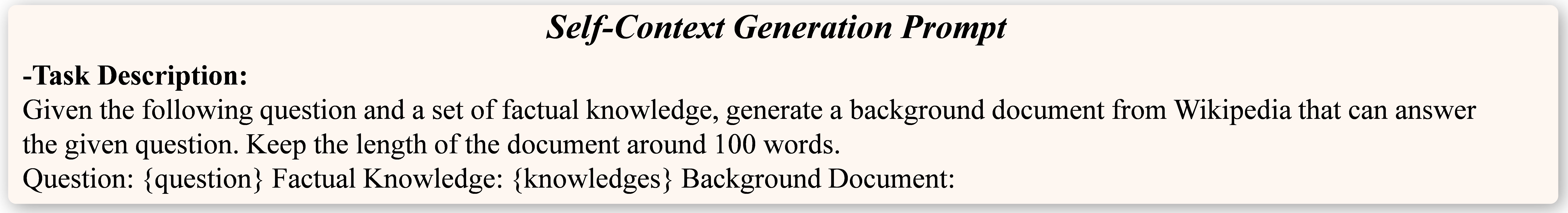}
    \label{fig:prompt2}
  \end{subfigure}
  
  \begin{subfigure}[b]{1.\textwidth}
    \centering
    \includegraphics[width=\textwidth]{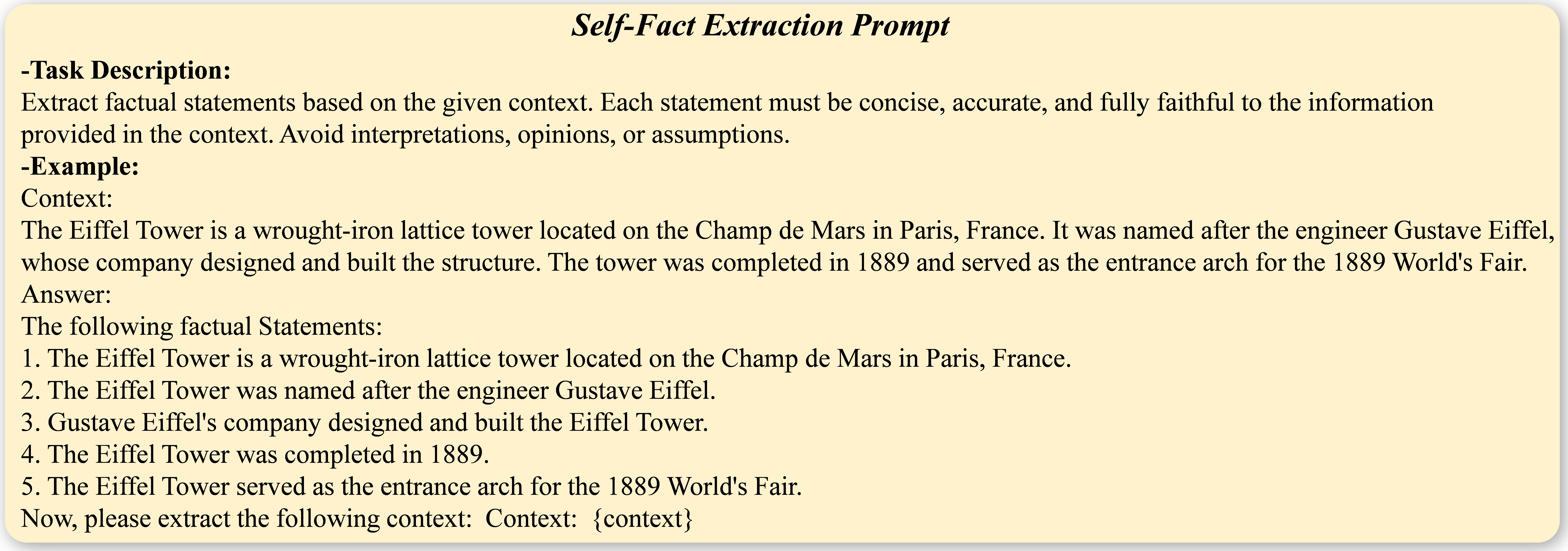}
    \label{fig:prompt3}
  \end{subfigure}
  
  \begin{subfigure}[b]{1.\textwidth}
    \centering
    \includegraphics[width=\textwidth]{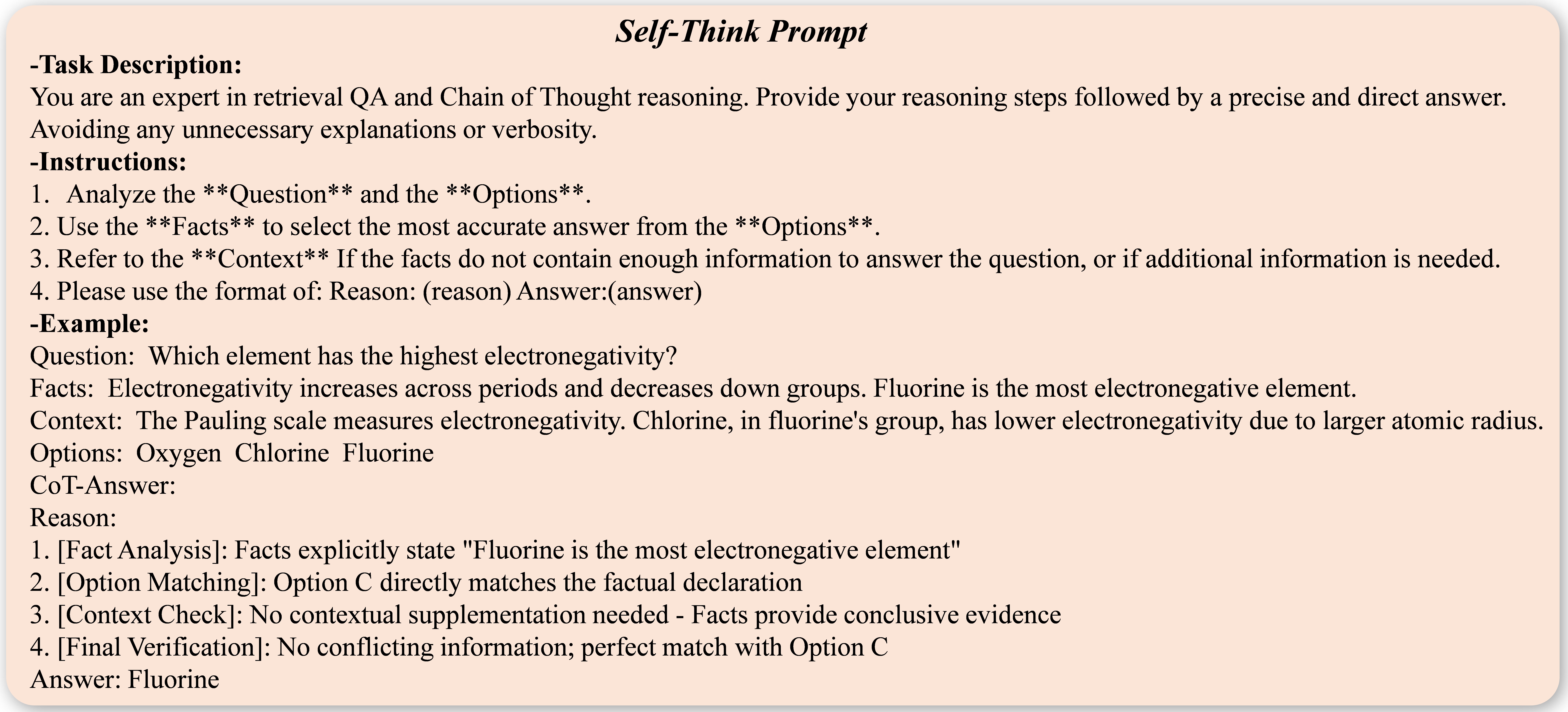}
    \label{fig:prompt4}
  \end{subfigure}
  \vspace{-3mm}
  \caption{Prompts for self-knowledge extraction, self-context generation, self-fact extraction, and the self-think.}
  \label{fig:prompts}
\end{figure*}

\clearpage

\noindent\textbf{Context-faithful Decoding:} These models~\cite{cad,coiced} modify the original inference strategy during model inference. Techniques such as contrastive decoding and constrained decoding are employed to guide the model to focus more on the given context rather than relying on parametric knowledge, thereby enhancing the model's context faithfulness.

\subsection{Evaluation Metrics and Implementation}\label{subsec:appendix-evaluation}

Our primary evaluation metric across all tasks is accuracy (ACC). We first normalize predictions and answers by removing stop words and punctuation, then determine whether the prediction and answer are identical. For the Memorization Ratio ($M_R$), following the approach~\cite{longpre-etal-2021-entity}, we first compute the Exact Match (EM) between predictions and original answers, denoted as $p_o$. Then, we compute the EM between predictions and substituted answers, denoted as $p_s$. Finally, we use the formula $M_R=\frac{p_o}{p_o+p_s}$. 

During inference (except for context-faithful decoding strategies), we deploy vLLM~\cite{kwon2023efficient}, a high-performance LLM designed to accelerate LLM inference. We standardized the sampling arguments across all methods and set the temperature to 0 to ensure reproducibility of results. 
To help the model better understand the task's patterns and requirements, all the baselines in our comparison adopt a few-shot format.

In the Contextual Knowledge Alignment module, we use the conventional Fixed-size Chunking Strategy to chunk the original context. Specifically, we divided the text into multiple segments based on the predefined fixed size, which facilitates the segmentation of the context for further processing. We set the default chunk size to 20. When selecting the top-k self-aligned context, we set $K$=5. In our paper, we did not report parameter analysis on $K$ and chunk size, as our model is not particularly sensitive to this hyperparameter, and the default settings already demonstrate the effectiveness of our approach.

\subsection{Implementation Details of Ablation Study}\label{subsec:appendix-generation}


To verify the effect of the Self-Think module, we conducted three separate ablation experiments. First, we conduct an ablation study on the overall module. Specifically, we directly prepend the self-aligned context to the original context to form a new combined context and instruct the LLM to answer based on it. This approach primarily leverages the Position Bias of the LLM's internal attention, which prioritizes earlier positions in the context~\cite{hsieh-etal-2024-found-in-the-middle,liu-etal-2024-lost}, thereby implicitly emphasizing the self-aligned context. The prompt is as follows:

\begin{tcolorbox}[
    colframe=black,        
    colback=gray!15,       
    coltitle=white,        
    fonttitle=\bfseries,   
    title=Variant 1: w/o whole Module  
]
 Context: \{self-aligned context\} \\
               \{origin context\} \\
    Question: \{question\} \\
    Answer: 
\end{tcolorbox}

Next, we conduct an ablation study on Think stage. Instead of the original Think stage, we introduce a Special Annotation approach, where we use special annotation to highlight the self-aligned context within the original context and employ instruction-based prompting to explicitly guide the LLM to focus on the self-aligned context enclosed by these markers. This modification prevents the LLM from actively thinking through and understanding the self-aligned facts, thereby hindering their effective fusion with the original context. The prompt is as follows:


\begin{tcolorbox}[
    colframe=black,        
    colback=gray!15,       
    coltitle=white,        
    fonttitle=\bfseries,   
    title=Variant 2: w/o Think  
]
- Instructions \\
     1. Analyze the \textbf{Context} and identify the sentences wrapped in '[important chunk: xxx]'. These sentences contain key information. \\
     2. Focus on the \textbf{important chunks} to extract the most relevant facts related to the **Question**. \\
     3. If the facts from the important chunks are not sufficient to answer the question, refer to the full \textbf{Context} for additional information. \\
     4. Please use the format of: Reason: (reason) Answer:(answer) \\
     Context: \{context\} \\
     Question: \{question\} \\
     Answer: 
\end{tcolorbox}

Finally, we conduct an ablation study on Reasoning stage. We substituted the structured reasoning with a naive Chain-of-Thought (CoT) approach, which does not provide explicit structured guidance for the LLM’s reasoning. This modification makes the model rely solely on its implicit inference capabilities. The prompt is as follows:

\begin{tcolorbox}[
    colframe=black,        
    colback=gray!15,       
    coltitle=white,        
    fonttitle=\bfseries,   
    title=Variant 3: w/o Reasoning  
]
- Instructions \\
     You are an expert in retrieval QA and Chain of Thought reasoning. Provide your reasoning steps followed by a precise and direct answer.Avoiding any unnecessary explanations or verbosity. Please use the format of: Reason: (reason) Answer:(answer) \\
     Context: \{context\} \\
     Question: \{question\} \\
     Answer: 
\end{tcolorbox}

\label{appendix:variant-self-think}

\section{Prompt Design}
The prompt templates applied in FaithfulRAG for self-knowledge extraction, self-context generation, self-fact extraction, and self-think are shown in the figure \ref{fig:prompts}.

\end{document}